\documentclass[letterpaper, 10 pt, conference]{ieeeconf}  

\IEEEoverridecommandlockouts                              

\usepackage{xcolor}
\usepackage{soul}
\usepackage{microtype}
\usepackage{graphicx}
\usepackage{fancyhdr}

\usepackage[a-2b,mathxmp]{pdfx}[2018/12/22]
\usepackage{colorprofiles}

\usepackage[noadjust]{cite}

\fancypagestyle{FirstPage}{

\lfoot{\footnotesize \copyright2023 IEEE. Personal use of this material is permitted. Permission from IEEE must be obtained for all other uses, in any current or future media, including reprinting/republishing this material for advertising or promotional purposes, creating new collective works, for resale or redistribution to servers or lists, or reuse of any copyrighted component of this work in other works.}

}

\title{\LARGE \bf
Evaluation of self-supervised pre-training for automatic infant movement classification using wearable movement sensors
}

\author{Einari Vaaras$^{1}$, Manu Airaksinen$^{2}$, Sampsa Vanhatalo$^{3}$, Okko Räsänen$^{1}$
\thanks{$^{1}$Einari Vaaras and Okko Räsänen are with the Unit of Computing Sciences, Tampere University, Finland. Email: \textit{einari.vaaras@tuni.fi}, \textit{okko.rasanen@tuni.fi}}
\thanks{$^{2}$Manu Airaksinen is with Helsinki University Hospital, Helsinki, Finland. Email: \textit{manu.airaksinen@hus.fi}}
\thanks{$^{3}$Sampsa Vanhatalo is with BABA Center, Pediatric Research Center, Department of Clinical Neurophysiology, New Children's Hospital and HUS Imaging, Helsinki University Hospital, Helsinki, Finland. Email: \textit{sampsa.vanhatalo@helsinki.fi}}
}

\begin{document}

\maketitle
\thispagestyle{empty}
\pagestyle{empty}

\begin{abstract}
The recently-developed infant wearable MAIJU provides a means to automatically evaluate infants' motor performance in an objective and scalable manner in out-of-hospital settings. This information could be used for developmental research and to support clinical decision-making, such as detection of developmental problems and guiding of their therapeutic interventions. MAIJU-based analyses rely fully on the classification of infant's posture and movement; it is hence essential to study ways to increase the accuracy of such classifications, aiming to increase the reliability and robustness of the automated analysis. Here, we investigated how self-supervised pre-training improves performance of the classifiers used for analyzing MAIJU recordings, and we studied whether performance of the classifier models is affected by context-selective quality-screening of pre-training data to exclude periods of little infant movement or with missing sensors. Our experiments show that i) pre-training the classifier with unlabeled data leads to a robust accuracy increase of subsequent classification models, and ii) selecting context-relevant pre-training data leads to substantial further improvements in the classifier performance.
\newline

\indent \textit{Clinical relevance}— This study showcases that self-supervised learning can be used to increase the accuracy of out-of-hospital evaluation of infants' motor abilities via smart wearables.
\end{abstract}

\section{INTRODUCTION} \label{sec_introduction}

Development of motor abilities in young infants (approx. 5--18 months) can be efficiently measured by the recently-developed MAIJU wearable (Motor ability Assessment of Infants with a JUmpsuit) \cite{maiju_scientific_reports, maiju_commsmed}, which applies human activity recognition (HAR) systems via smart wearables to clinical use. MAIJU provides an objective and scalable method for automatic evaluation of infants' motor abilities that can be used in a home environment. This information could be used for developmental research as well as to support clinical decision-making in situations like early detection of cerebral palsy or guidance of therapeutic interventions \cite{maiju_scientific_reports, maiju_commsmed}. As shown in \cite{maiju_commsmed}, the accuracy of analysis produced by MAIJU is comparable to that of human evaluation from a concurrent video recording. MAIJU analysis relies on the categorical classification of infant posture, movement, and carrying, and the reliability of the final analyses depend on the accuracy of the classification tasks. Therefore, it is essential to study ways to increase the accuracy of these classifications in order to build more robust and reliable automatic evaluation of infants' motor abilities using movement detection, such as is present in MAIJU.

When building machine-learning models, a common scenario is that there is a scarcity of labeled data but an abundance of unlabeled data \cite{state_of_ssl_har_using_wearables, selfhar, sense_and_learn, ssl_sorting_sequences, wav2vec2, hubert}. In self-supervised learning (SSL), a machine-learning model creates representations of input data without relying on human-annotated labels. Recently, the use of SSL for pre-training models has shown great success in various fields, such as speech processing \cite{cpc, wav2vec2, hubert, data2vec_paper, data2vec2}, computer vision \cite{cpc, cpc_image_recognition, byol, mae, data2vec_paper, data2vec2, ssl_sorting_sequences, contrastive_learning_of_visual_representations}, and natural language processing \cite{cpc, bert, data2vec_paper, data2vec2}.

SSL has also been successfully applied to HAR. Tang et al. \cite{selfhar} presented a semi-supervised method which utilizes unlabeled data via SSL to promote the performance of HAR systems. Their method achieved state-of-the-art performance with multiple HAR datasets, and the method matched the performance of similar supervised learning-based approaches with up to 10 times less labeled training data. Saeed et al. \cite{sense_and_learn} proposed an SSL framework consisting of multiple auxiliary tasks for learning useful features from unlabeled raw sensory data. Their experiments demonstrated that the proposed method obtained competitive results with supervised learning-based approaches with as few as five labeled training examples for each target class.

In the present study, we investigate if SSL pre-training improves the performance of automatic infant motility assessment using the MAIJU wearable. For this, we use the classification of movement for our MAIJU data as a test case as it was the most difficult classification task in \cite{maiju_scientific_reports} and \cite{maiju_commsmed}. While doing so, we also seek an answer to whether classification model performance is improved by context-selective quality-screening of the data used for pre-training. The latter question is important because long-term movement recordings may include substantial amounts of data from periods that are not informative for assessing infant's own motor performance, such as sleeping or carrying.

\thispagestyle{FirstPage}

\section{METHODS} \label{sec_methods}

\begin{figure}[t]
    \centering
    \vspace{8 pt}
    \includegraphics[width=0.48\textwidth]{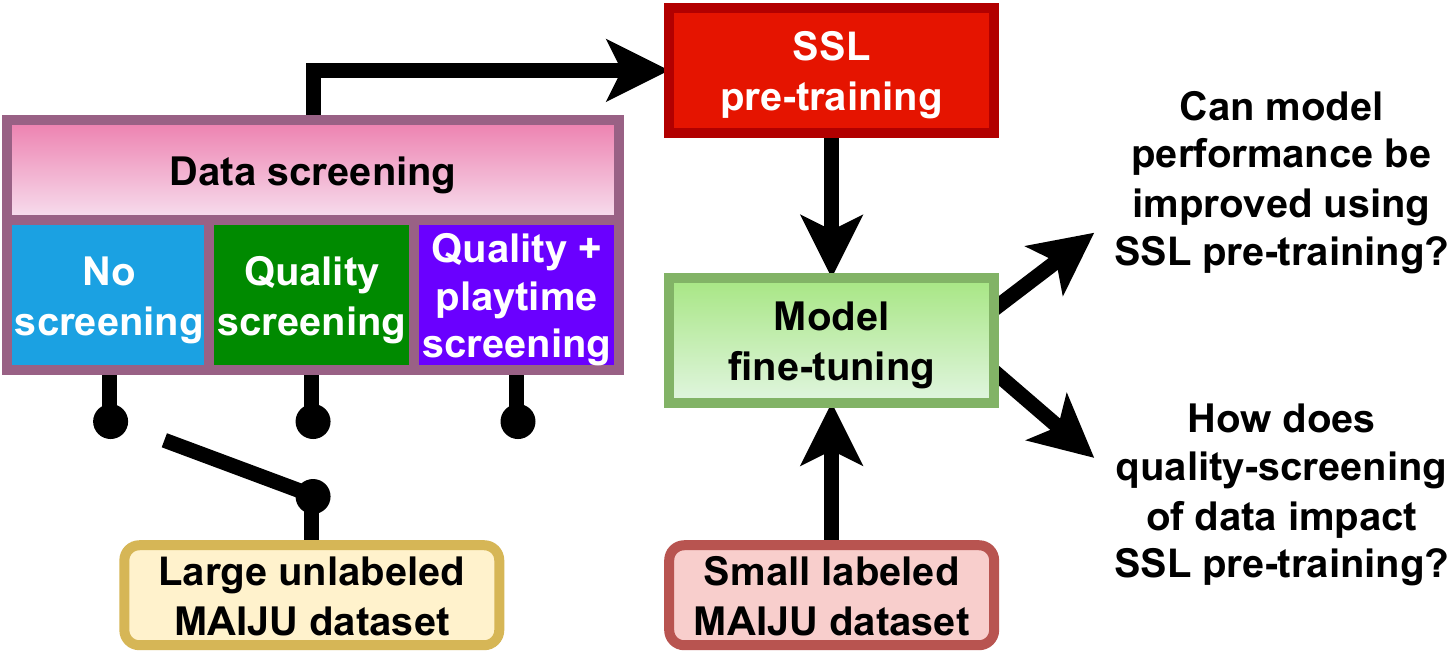}
    \caption{An overview of the present experimental setup. Unlabeled MAIJU data is first used as input to SSL pre-training using three alternative data screening conditions, followed by fine-tuning the pre-trained models using labeled MAIJU data.}
    \label{fig_block_diagram}
\end{figure}

Fig. \ref{fig_block_diagram} shows an overview of the present experiments. First, a large unlabeled dataset of MAIJU data (see Sec. \ref{subsec_data}) is used as input to SSL pre-training using three alternative data screening conditions. For SSL, we use the recently proposed data2vec \cite{data2vec_paper} algorithm. After the pre-training phase, we utilize a smaller labeled dataset of MAIJU data (Sec. \ref{subsec_data}) to fine-tune the pre-trained models. Finally, fine-tuned models are evaluated to find out if we can improve model performance using SSL pre-training, and whether quality-screening of unlabeled data affects the benefits of SSL pre-training.

\subsection{Self-supervised pre-training with data2vec} \label{subsec_data2vec}

Data2vec \cite{data2vec_paper} is an SSL algorithm designed to work with multiple different data modalities. To achieve this, the method combines masking and predicting its own latent feature representations. As argued in \cite{data2vec_paper}, the power of the method comes from contextualized target representations produced by the self-attention mechanism in Transformer models. Although data2vec has been proven to work well with multiple data modalities \cite{data2vec_paper, data2vec2, data2vec_alzheimer}, it should be noted that the method is not entirely data-agnostic since the method uses modality-specific feature encoders.

The data2vec method consists of two neural networks, a student network and a teacher network. Both networks have the same architecture, and they consist of a feature encoder followed by a Transformer network consisting of $L$ Transformer encoder blocks. For the student network, the feature embeddings produced by the encoder are partially masked, whereas the teacher network receives unmasked embeddings. The student model then tries to predict the content of the masked sections of the embeddings based on the unmasked sections. The training targets are the averaged outputs of the top $K$ Transformer encoder blocks of the teacher network, whereas the model predictions are the outputs of the last student network layer. As proposed by \cite{data2vec_paper}, the training targets are normalized before averaging in order to prevent representation collapse (i.e. the encoder outputs a constant time-invariant feature representation) and to prevent Transformer blocks with large output values to dominate the target features.

The weights of the teacher network are an exponentially moving average (EMA) \cite{ema_original} of the student network weights, i.e. after each minibatch update of the student network

\begin{equation} \label{info_nce}
    \theta_t = \tau \theta_t + (1 - \tau) \theta_s \ ,
\end{equation}
where $\theta_t$ are the weights of the teacher network, $\theta_s$ are the weights of the student network, and $\tau$ is a hyperparameter controlling the rate at which the weights of the teacher network are updated. The value of $\tau$ changes during training from a value of $\tau_0$ into $\tau_{end}$ linearly in $\tau_n$ minibatch updates.

\section{EXPERIMENTS} \label{sec_experiments}

\subsection{Data} \label{subsec_data}

We conducted our pre-training experiments using a large unlabeled dataset of MAIJU data (172 recordings, 62 distinct participants, 1333 h, referred to as ``DS1" from now on), and for our fine-tuning experiments we utilized a smaller labeled dataset of MAIJU data \cite{maiju_commsmed} (41 recordings and distinct participants, 29.3 h, referred to as ``DS2"). Most of the recordings come from typically developing infants. There are 11 recordings (3 participants) in DS1 and 3 recordings (3 participants) in DS2 with diagnosed motor disabilities. For both datasets, each sensor from all four limbs produced data in 3 accelerometer and 3 gyroscope channels (x, y, and z) with a sampling frequency of 52 Hz, resulting in 24-channel data. Following \cite{maiju_commsmed, maiju_model_comparison, maiju_scientific_reports}, the raw signals were pre-processed by interpolating the data packets received from each sensor (via a Bluetooth connection) into a common 52-Hz time-stamp base, removing gyroscope bias, and applying temporal smoothing with a seven-tap median filter. The pre-processed raw signals were then windowed into 120-sample frames (2.3 seconds) with an overlap of 50\%.

DS1 contains multiple-hour to day-long recordings of 4--20 month-old infants obtained from their homes with a minimally controlled recording setup. In addition, the data comes with parental reports of which segments of the long recording consisted of free-play (autonomous active movement) of the baby, here referred to as ``playtime". Since DS1 was gathered in a minimally-controlled manner, there is no guarantee that the baby has been awake or let alone whether the MAIJU jumpsuit has been on the baby outside of the reported playtime. Furthermore, since the raw sensor data has been streamed over a Bluetooth connection, it is highly probable that one or multiple sensors lose connection occasionally. Consequently, without any context-selective quality-screening, a high proportion of the data is so-called ``flat data" in which there is very little signal variation over time. This flat data is not informative for assessing infants' motor performance, and therefore it is relevant to examine whether including or discarding the flat data in DS1 via quality-screening affects pre-training, as described in Sec. \ref{subsec_pretraining}.

\begin{table}[t]
    \centering
    \vspace{8 pt}
    \caption{Annotated movement categories and their frame counts for DS2 dataset used for model fine-tuning and testing. For each movement category, the number of frames, the total duration in minutes, and the percentage of frames compared to the total number of categorized data is given. The number of frames which are not qualified for model fine-tuning and testing (baby out of screen or being handled by caregiver) is also reported.}
    \includegraphics[width=0.40\textwidth]{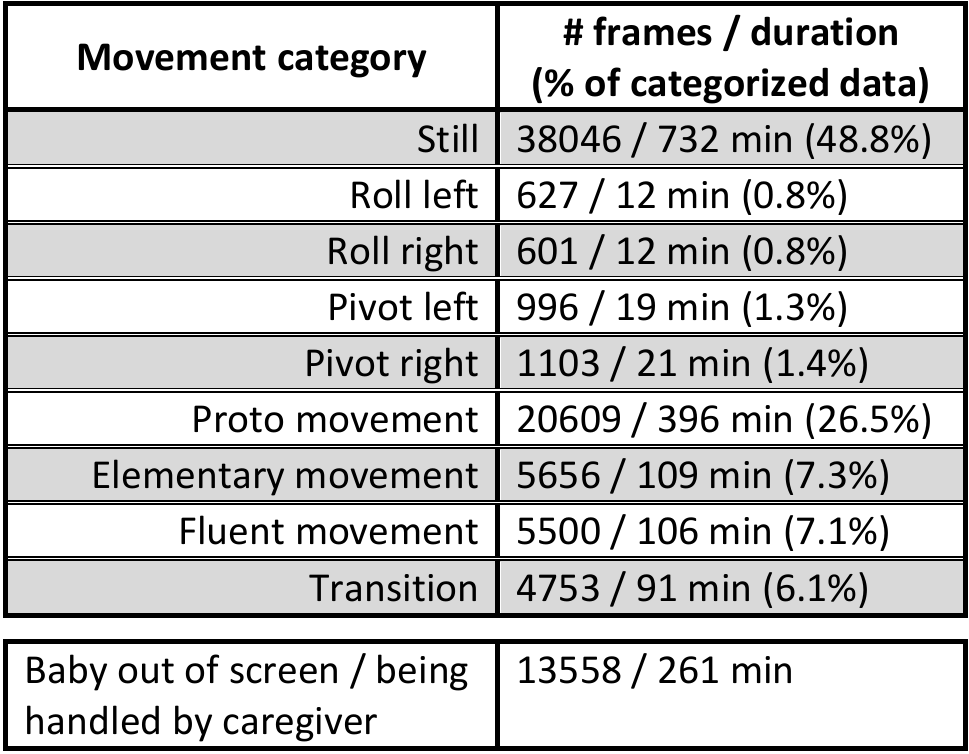}
    \label{table_labeled_dataset_statistics}
\end{table}

DS2 contains up to one hour of data per recording, and it was obtained within a semi-controlled recording setup with parallel video at the infants' homes ($N = 17$ recordings) or at the lab in a home-like setting ($N = 24$). The data contains nine annotated movement categories for each 2.3-s frame: still, roll left, roll right, pivot left, pivot right, proto movement, elementary movement, fluent movement, and transition. Each movement category was annotated by either two ($N = 9$) or three ($N = 32$) domain experts, and the data was quality-screened by the annotators during the annotation process. Table \ref{table_labeled_dataset_statistics} gives further details on the number of frames and the total duration in minutes for each annotated movement category. We use all labeled  data for the model training process, but for testing our trained models we only use the frames in which the three annotators agreed on the label. For training our models, we used the so-called iterative annotation refinement (IAR) labels from \cite{maiju_commsmed}. The use of IAR labels was shown in \cite{maiju_scientific_reports} to yield better classification performance compared to the original annotator ratings.

\subsection{Model pre-training} \label{subsec_pretraining}

For data2vec pre-training, the feature encoder was the same convolutional neural network (CNN) sensor encoder module that was used in \cite{maiju_commsmed}. The module combines the raw accelerometer and gyroscope signals at a frame-level and outputs 160-dimensional latent representations for each input frame. The only modification to the CNN encoder was the addition of layer normalization \cite{layernorm_original} after the last two convolution operations in order to make the pre-training process more stable. The Transformer model consisted of $L = 12$ Transformer encoder blocks with an input and output dimensionality of 160, a feed-forward inner dimensionality of 640, and 10 attention heads in each multi-head self-attention. As in the original data2vec implementation \cite{data2vec_paper}, a Gaussian error linear unit (GeLU) \cite{gelu_original} activation function was used in the Transformer model, and a linear projection layer consisting of 160 units was added both after the encoder and the Transformer. Instead of using absolute positional encodings with sinusoids (as in e.g. \cite{attention_is_all_you_need}), a relative positional encoder was used in the present experiments. Similar to e.g. \cite{wav2vec2, data2vec_paper}, this relative positional encoder was a 1D convolution (kernel size = 13, padding = 6, stride = 1) followed by a GeLU activation and layer normalization.

For data2vec pre-training, DS1 was randomly split into a training and validation set in a ratio of 80:20 sequences of 260 consecutive frames each (i.e. 5 minutes). End-of-recording sequences shorter than 260 frames were padded into a length of 260 frames, and these padded frames were then ignored by the data2vec model during pre-training. Pre-training was tested using three data screening conditions:
\begin{enumerate}
    \item \textit{No screening}: Utilizing all unlabeled data (16,081 sequences, 1333 hours).
    \item \textit{Quality screening}: Screening out all the sequences in which over 20\% of the frames have more than one sensor dropped out (13,119 sequences, 1087 hours).
    \item \textit{Quality + playtime screening}: Screening out all the sequences in which over 20\% of the frames have more than one sensor dropped out and/or the data was not marked as playtime (4,669 sequences, 387 hours).
\end{enumerate}
Each minibatch consisted of 32 sequences (160 minutes of signal). Mean squared error loss and Adam \cite{adam_original} optimizer were used. The loss was only computed for the masked frames. As training targets, we use the instance-normalized \cite{instancenorm_original} and averaged outputs of all Transformer encoder blocks ($K = 12$) of the teacher model. Similar to \cite{data2vec_paper}, we use the outputs of the feed-forward parts of each Transformer block as the outputs of our teacher model. Early stopping with a patience of 25 epochs based on the validation loss was used, and the weights of the best student model according to the validation loss were used for fine-tuning (Sec. \ref{subsec_finetuning}). An initial learning rate of $10^{-4}$ was used with a reduction factor of 0.5 based on the validation loss with a patience of 15 epochs.

For masking the embeddings of the student network, we randomly select frames in each minibatch with a probability of $p = 0.15$ to be mask starting indices. For each mask starting index we mask the embedding of that frame and the next two frames, resulting in a minimum mask length of three frames. Mask spans can overlap, which results in approximately 38\% of each sequence to be masked on average. Contrary to \cite{data2vec_paper}, we do not replace the masked embeddings with a learnable mask token, but instead we simply replace the masked embeddings with zeros. For the EMA update of the teacher weights, we use $\tau_0 = 0.9998$, $\tau_{end} = 0.99999$, and $\tau_n = 10,000$. As in \cite{data2vec_paper}, the weights of the feature encoder and the positional encoder were shared between the student and teacher networks, and therefore the EMA update only concerned the Transformer part of the network.

Each pre-training experiment was run both with and without data augmentation (see \cite{maiju_model_comparison} for further details on the data augmentation methods). The augmentation methods that were used were sample dropout ($p = 0.1$), sensor dropout ($p = 0.1$), and sensor rotation ($p = 0.3$). For sensor rotation, we utilized the Euler rotation formula where the yaw, pitch, and roll angles were sampled randomly from a uniform distribution between $\pm15^{\circ}$. Data augmentation was only applied to the training samples. As noted in e.g. \cite{contrastive_learning_of_visual_representations}, the use of data augmentations may play a crucial role in SSL pre-training.

During pilot experiments, we noticed that representation collapse was very likely to occur during pre-training when a high proportion of input data was flat. These cases were when playtime screening was not utilized and when data augmentation was used with a too high probability for sample or sensor dropout. Hence, for our pre-training experiments, sample and sensor dropout probabilities were kept at a relatively low value ($p = 0.1$) when data augmentation was used, and pre-training was restarted if a representation collapse occurred.

\subsection{Model fine-tuning with movement category labels} \label{subsec_finetuning}

DS2 was used for fine-tuning the pre-trained models (Sec. \ref{subsec_pretraining}) for the infant movement classification task. The models were trained using recording-level 10-fold cross-validation in which each training fold was split on a recording-level so that approximately 80\% of the training fold recordings were used for training and the rest for validation. The classification model produces a prediction of the infant's movement category for every input frame. Each training was run three times, once with a full training and validation set, and also with randomly selecting 5\% or 15\% of the training and validation set samples to simulate having smaller labeled datasets. This information could be utilized when building classifiers from scratch for other similar domains. At test time, an aggregate confusion matrix was calculated across the folds, followed by unweighted average F1 (UAF1) score calculation from the matrix.

The pre-trained models were fine-tuned in two stages. In the first stage, two randomly initialized fully-connected exponential linear unit (ELU) \cite{elu_original} layers followed by a softmax function were added after the Transformer encoder blocks in order to turn the Transformer output into categorical probabilities for each movement category. These two ELU layers consisted of 160 and 9 units, with a dropout of 40\% after the first layer. These layers were fine-tuned separately by freezing the weights of the encoder and Transformer parts of the network. Weighted categorical cross-entropy loss, Adam optimizer, and a batch size of 260 consecutive frames were used. Substantial class imbalance was handled during training by weighting classes in loss calculation by their inverse frequency. Early stopping with a patience of 100 epochs based on the UAF1 score of the validation set was used to select the best-performing model. An initial learning rate of $4\cdot10^{-5}$ was used with a reduction factor of 0.5 based on the UAF1 score of the validation set with a patience of 30 epochs.

In the second stage, the entire network was fine-tuned with the same training hyperparameters as in the first stage of the training process with three exceptions: First, at the beginning of the training, the learning rate was linearly increased from $4\cdot10^{-8}$ to $4\cdot10^{-5}$ in 20 epochs, after which the same learning rate scheduler was used as in the first stage. Second, to match the model training in \cite{maiju_commsmed}, sample dropout ($p = 0.3$) and sensor dropout ($p = 0.3$) were applied randomly to the training samples as a means of data augmentation in order to make the trained models more robust. Third, a dropout of 40\% was used in the Transformer part of the network, both after the self-attention blocks and also after both fully-connected layers in the feed-forward blocks.

During our experiments, we observed that by adding randomly-initialized classification layers to the end of the network and then fine-tuning the entire model at once nullifies the benefit of pre-training. We also observed that the same adverse effect occurs if the learning rate is too high at the start of the second fine-tuning stage.

\subsection{Classification baseline} \label{subsec_baseline}

As a baseline for our experiments, we used the same WaveNet-based \cite{wavenet} classification model that was used in \cite{maiju_commsmed, maiju_scientific_reports, maiju_model_comparison}. WaveNet is a CNN-based model that combines causal filters with gated dilated convolutions, and it has been shown to work particularly well with MAIJU data (see e.g. \cite{maiju_model_comparison}). Furthermore, in order to better compare the benefit of pre-training, we include a non-pretrained (randomly initialized) Transformer model baseline in our experiments. This Transformer model has the same architecture as the fine-tuned data2vec Transformer models in Sec. \ref{subsec_finetuning}. Both the WaveNet and the Transformer baselines were directly trained with DS2 using the same training hyperparameters as in the second pre-training stage, with the exception of not having the learning rate linearly increased at the beginning of training. Since the data2vec method requires having a Transformer-based architecture, we do not experiment with pre-training the WaveNet model. However, it should be noted that there are also SSL methods that could be used to pre-train the WaveNet model, such as \cite{cpc} and \cite{byol}.

\section{RESULTS} \label{sec_results}

\begin{table}[t]
    \centering
    \vspace{8 pt}
    \caption{The results of the present experiments. The highest UAF1 (\%) score for each training data sampling size is \textbf{bolded}. The results with an UAF1 score lower than the WaveNet baseline are marked as \textcolor{red}{red}, and the results with an UAF1 score higher than the baseline are marked as \textcolor{blue}{blue}.}
    \includegraphics[width=0.48\textwidth]{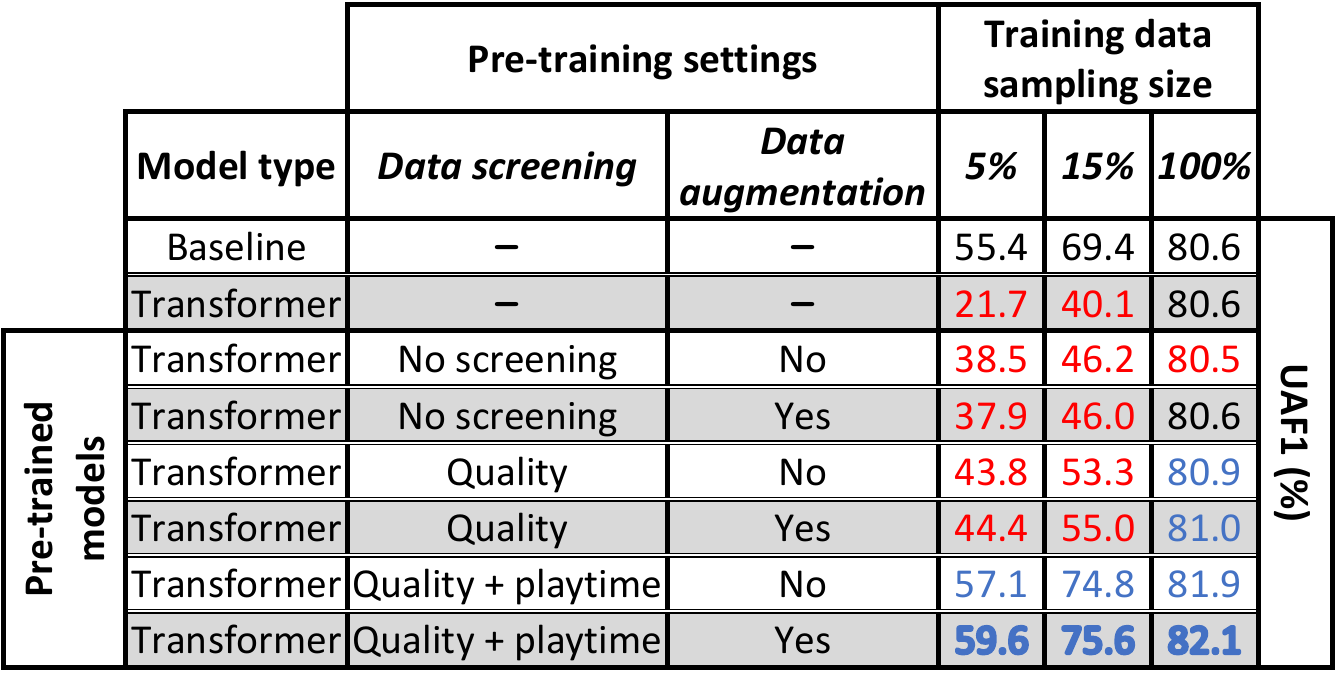}
    \label{table_results}
\end{table}

\begin{figure}[t]
    \centering
    \vspace{8 pt}
    \includegraphics[width=0.43\textwidth]{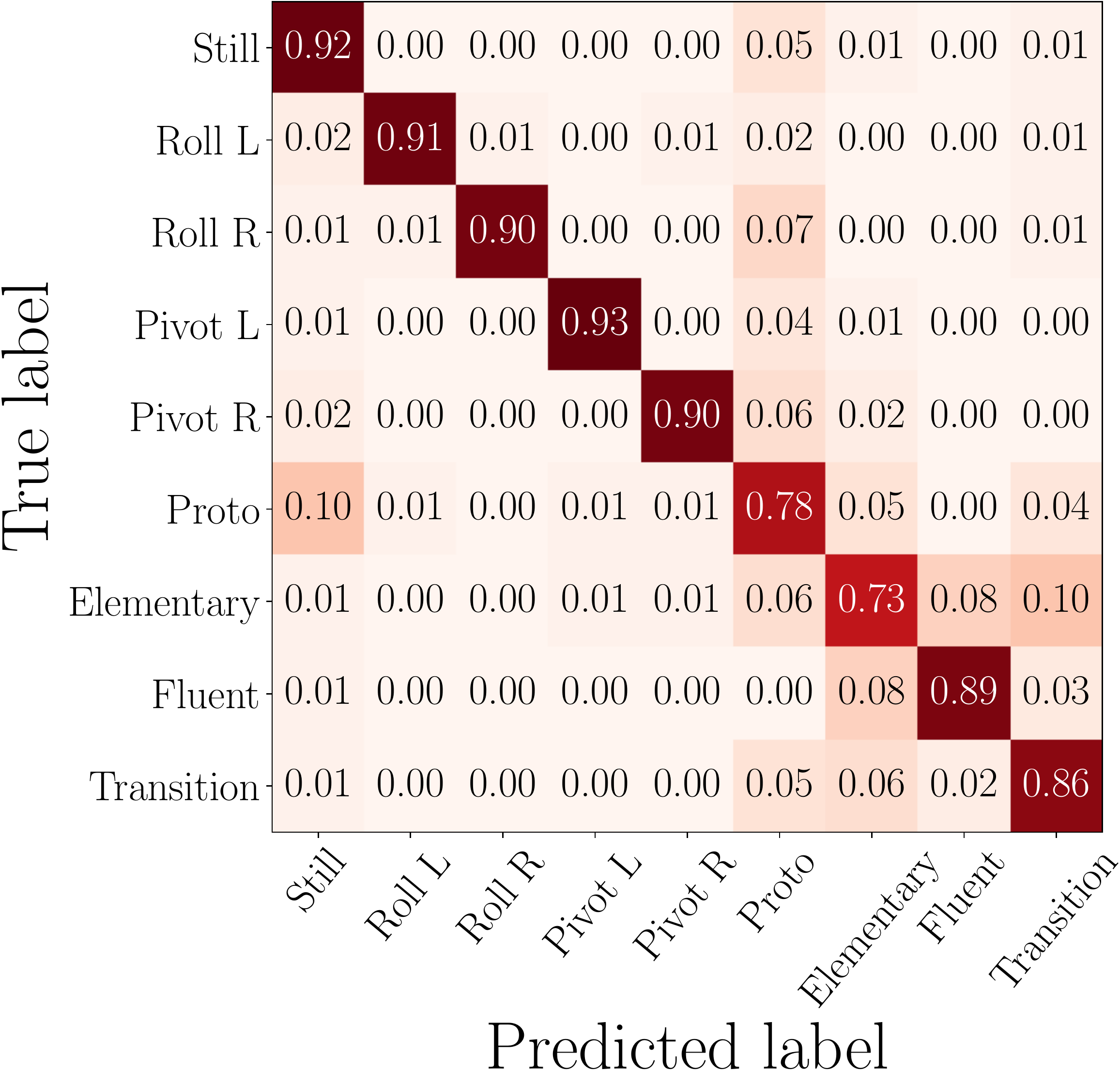}
    \caption{The confusion matrix normalized for recall values for the best-performing classification model, computed as an aggregate confusion matrix across all folds in the recording-level 10-fold cross-validation.}
    \label{fig_conf_mat}
\end{figure}

Results of the experiments are presented in Table \ref{table_results}. The results demonstrate that overall the baseline model, WaveNet, is very competitive performance-wise compared to the Transformer model, even though the Transformer model has nearly 34 times the number of trainable parameters of the WaveNet model. With training data sampling sizes of 5\% and 15\%, it is evident that the number of training samples is too small for the randomly initialized Transformer model since the model's performance is notably below that of the WaveNet model. In these cases the WaveNet model excels, most probably due to the fact that it does not overfit to the training data as easily as the Transformer model when the number of training samples is small and when the models do not have any a priori knowledge of the data. However, utilizing pre-training without any data screening or with quality screening gives a large performance boost compared to using a non-pretrained model, albeit this increase in performance does not make the Transformer models outperform the WaveNet model which did not utilize pre-training. Only when using the most high-quality data for pre-training with ``quality + playtime" screening does the Transformer model outperform the WaveNet model.

When using all labeled data for training (100\% sampling size), there was no practical difference between the performance of the WaveNet model, the randomly initialized Transformer model, and the pre-trained model when no data screening was used during pre-training. Using quality screening gave a small performance boost of 0.4 percentage points in terms of UAF1 score, while using ``quality + playtime" screening gave the best overall performance with a UAF1 score of 82.1\%. Data augmentation during pre-training helped in achieving a better performance level, but only when data screening was used. Overall, the results indicate that pre-training is essential for models with a large number of trainable parameters, in particular if the number of labeled training samples is small. The results also show that data quality screening plays a crucial role in pre-training since a better classification performance was always achieved with it, other conditions being equal.

The aggregate confusion matrix for the best-performing classification model (pre-trained Transformer with ``quality + playtime" screening and data augmentation) is shown in Fig. \ref{fig_conf_mat}. Overall, the classifier performs well across all nine movement categories, as denoted by the high recall values on the diagonal of the matrix. The most difficult categories to predict were ``proto movement" and ``elementary movement", for which the former got often misclassified as ``still" and the latter as either ``fluent movement" or ``transition". The corresponding mean precision and recall, as derived from the confusion matrix, were 78.3\% and 86.9\%, respectively.

One potential explaining factor for the differences observed due to data screening could be attributed to the variance of the embeddings produced by the feature encoder. We observed that when less data screening was used, the proportion of flat data was higher in the pre-training dataset. This, on the other hand, caused the feature encoder to learn a feature representation with very little variance over time, and consequently, the Transformer model has an easier task in predicting masked latent feature representations in this low-variance feature space. In contrast, we observed that with more data screening, the feature encoder learned a richer feature representation in terms of variance over time, which also means that the Transformer model has to solve a more complex task when it tries to predict masked latent feature representations in a higher-variance feature space. An additional cause might simply be that with more data screening, the pre-training data represents the labeled fine-tuning data more closely.

\section{CONCLUSIONS} \label{sec_conclusions}

In the present study, we investigated whether SSL pre-training can improve the performance of automatic assessment of infants' movement from wearable recordings with the MAIJU jumpsuit; additionally, we assessed how much context-selective quality-screening of pre-training data may improve model fine-tuning performance. Our results show that pre-training is crucial for models with a large number of trainable parameters, especially if the amount of labeled training data is small. Our results also show that screening of pre-training data is important and it leads to substantial increase in model performance; the effect is particularly clear when the pre-training data is selected to only include data with direct relevance for the eventual classifier tasks. These findings could be used to further encourage research of SSL methods for HAR and clinical healthcare applications. Potential future work includes further developing multimodal SSL algorithms, such as the data2vec algorithm used in the present experiments, to be more robust against representation collapse during SSL pre-training.

\addtolength{\textheight}{-12cm}   




\section*{ACKNOWLEDGMENT}

O.R. was supported by Academy of Finland (AKA) grants no. 345365 and 314602. E.V. was supported by AKA grant no. 335872 and a grant from Finnish Brain Foundation. M.A. was supported by AKA grants no. 314573, 335872, and 343498. S.V. was supported by AKA grants no. 314602, 335788, 335872, 332017, and 343498, Finnish Pediatric Foundation (Lastentautiensäätiö), Finnish Brain Foundation, Sigrid Juselius Foundation, and HUS Children’s Hospital/HUS diagnostic center research funds.

\bibliographystyle{IEEEtran}

\addtolength{\textheight}{2.2cm} 

\bibliography{mybib}

\end{document}